\documentclass{article}

\usepackage{microtype}
\usepackage{graphicx}
\usepackage{subfigure}
\usepackage{booktabs} 

\usepackage{hyperref}

\usepackage[accepted]{icml2025}

\usepackage{amsmath}
\usepackage{amssymb}
\usepackage{mathtools}
\usepackage{amsthm}

\usepackage[capitalize,noabbrev]{cleveref}

\theoremstyle{plain}
\newtheorem{theorem}{Theorem}[section]

\theoremstyle{definition}

\newtheorem{example}{Example}[section]
\theoremstyle{remark}

\usepackage{tikz}
\usetikzlibrary{automata,arrows.meta,calc,positioning,decorations.pathreplacing,backgrounds,matrix,arrows,patterns}
\usepackage{bm}

\usepackage[textsize=tiny]{todonotes}

\newcommand{\brac}[1]{\left( #1 \right)}
\newcommand{\sbrac}[1]{\left[ #1 \right]}
\newcommand{\set}[1]{\left\{ #1 \right\}}
\icmltitlerunning{Learning Structured Representations of LTL Instructions for Multi-Task RL}

\newcommand*{\given}{\,|\,}

\newcommand{\always}{\mathsf{G}\,}
\newcommand{\event}{\mathsf{F}\,}

\newcommand{\nex}{\mathsf{X}\,}
\newcommand{\until}{\;\mathsf{U}\;}

\newcommand{\gror}{\;|\;}

\DeclareMathOperator*{\argmax}{arg\,max}

\DeclareMathOperator*{\E}{\mathbb E}

\DeclarePairedDelimiter\abs{\lvert}{\rvert}%
\DeclarePairedDelimiter\norm{\lVert}{\rVert}%
\makeatletter
\let\oldabs\abs
\def\abs{\@ifstar{\oldabs}{\oldabs*}}
\let\oldnorm\norm
\def\norm{\@ifstar{\oldnorm}{\oldnorm*}}
\makeatother

\usepackage{pifont}

\usepackage{algorithm}
\usepackage{algpseudocode}

\begin{document}

\twocolumn[
\icmltitle{Zero-Shot Instruction Following in RL via Structured LTL Representations}



\icmlsetsymbol{equal}{*}

\begin{icmlauthorlist}
\icmlauthor{Mattia Giuri}{equal,ox}
\icmlauthor{Mathias Jackermeier}{equal,ox}
\icmlauthor{Alessandro Abate}{ox}
\end{icmlauthorlist}

\icmlaffiliation{ox}{Department of Computer Science, University of Oxford}

\icmlcorrespondingauthor{Mathias Jackermeier}{mathias.jackermeier@cs.ox.ac.uk}

\icmlkeywords{Machine Learning, ICML}

\vskip 0.3in
]



\printAffiliationsAndNotice{\icmlEqualContribution} 

\begin{abstract}
Linear temporal logic (LTL) is a compelling framework for specifying complex, structured tasks for reinforcement learning (RL) agents. Recent work has shown that interpreting LTL instructions as finite automata, which can be seen as high-level programs monitoring task progress, enables learning a single generalist policy capable of executing arbitrary instructions at test time. However, existing approaches fall short in environments where multiple high-level events (i.e., atomic propositions) can be true at the same time and potentially interact in complicated ways. In this work, we propose a novel approach to learning a multi-task policy for following arbitrary LTL instructions that addresses this shortcoming. Our method conditions the policy on sequences of simple Boolean formulae, which directly align with transitions in the automaton, and are encoded via a graph neural network (GNN) to yield structured task representations. Experiments in a complex chess-based environment demonstrate the advantages of our approach.
\end{abstract}

\section{Introduction}
In recent years, we have seen remarkable progress in training artificial intelligence (AI) agents to follow arbitrary instructions~\citep{luketina2019survey,liu2022GoalConditioned,paglieri2025BALROG,klissarov2025MaestroMotif}. One of the central considerations in this line of work is which type of instruction should be provided to the agent; while many works focus on tasks expressed in natural language~\citep{goyal2019Using,hill2020Human,carta2023Grounding}, there recently has been increased interest in training agents to follow instructions specified in \textit{formal} language~\citep{jothimurugan2021Compositional,vaezipoor2021LTL2Action,qiu2023Instructing,yalcinkaya2024Compositional,jackermeier2025deepltl}. As a type of \textit{programmatic task representation}, formal specification languages offer several desirable properties, such as well-defined semantics and compositionality. This makes formal instructions especially appealing in safety-critical settings, in which we want to define precise tasks with a well-defined meaning, rather than giving ambiguous natural language commands to the agent~\citep{leon2021Systematic}.

One particular formal language that has proven to be a powerful and expressive tool for specifying tasks in reinforcement learning (RL) settings is \textit{linear temporal logic} (LTL;~\citealp{pnueli1977temporal})~\citep{hasanbeig2018LogicallyConstrained,hahn2019OmegaRegular,kuo2020Encoding,vaezipoor2021LTL2Action,leon2022Nutshell,liu2024Skill}. LTL instructions are defined over a set of \textit{atomic propositions}, corresponding to high-level events that can hold true or false at each state of the environment. These atomic propositions are combined using logical and temporal operators, which allow for the specification of complex, non-Markovian behavior in a compositional manner, naturally incorporating aspects like safety constraints and long-term goals. Recent work has exploited the connection between LTL and corresponding automata structures (typically variants of B\"uchi automata;~\citealp{buchi1966Symposium}), which provide a \textit{programmatic} way to monitor task progress, to train generalist policies capable of executing arbitrary LTL instructions at test time~\citep{qiu2023Instructing,jackermeier2025deepltl}.

However, existing approaches often struggle in scenarios where multiple atomic propositions can hold true simultaneously. This is due to the fact that current methods treat possible assignments of propositions \textit{in isolation}, and do not explicitly model the complex interactions that may occur between different high-level events. In this paper, we develop a novel approach that addresses these limitations. Our approach translates transitions in the automaton into equivalent Boolean formulae, which provide succinct, structured representations of the propositions that are relevant for making progress towards the given task, and explicitly capture the logical conditions for the transition. We show that encoding these formulae via a graph neural network (GNN) yields meaningful representations that can be used to train a policy conditioned on different ways of achieving a given task, enabling zero-shot generalization to novel LTL instructions at test time.
Our main contributions are as follows:
\begin{itemize}
    \item we develop a novel approach to learning policies for following arbitrary LTL instructions that can effectively handle complex interactions between atomic propositions;
    \item we propose representing LTL instructions as sequences of Boolean formulae, and show how existing policy learning approaches can be augmented with GNNs to improve representation learning;
    \item we introduce a novel, chess-like environment where many different propositions can be true simultaneously, which allows us to study the performance of existing approaches in this challenging setting;
    \item lastly, we conduct an extensive empirical evaluation of our proposed method on this challenging environment, and show that it achieves state-of-the art results and outperforms existing methods.
\end{itemize}

\section{Related Work}
While significant research effort has explored the use of LTL to specify tasks in RL~\citep{fu2014Probably,degiacomo2018Reinforcement,hasanbeig2018LogicallyConstrained,camacho2019LTL,hahn2019OmegaRegular,bozkurt2020Control,cai2021Reinforcement,shao2023Sample,voloshin2023Eventual,le2024Reinforcement,shah2025LTLConstrained}, most approaches are limited to training agents to satisfy a \textit{single} specification that is fixed throughout training and evaluation. In contrast, we aim to learn a general policy that can zero-shot execute arbitrary LTL instructions at test time.

Several works have begun to tackle this challenge of learning generalist policies. Early methods, such as that of \citet{kuo2020Encoding}, propose composing recurrent neural networks (RNNs) that mirror the structure of LTL formulae, but this approach requires learning a non-stationary policy, which is generally challenging~\citep{vaezipoor2021LTL2Action}. Other approaches decompose LTL tasks into subtasks, which are then completed sequentially by a goal-conditioned policy \citep{araki2021Logical,leon2022Nutshell,liu2024Skill,xu2024Generalizationa}. However, such methods can exhibit myopic behavior, leading to globally suboptimal solutions because they do not consider the full specification structure during subtask execution. In contrast, our method conditions the policy on the entire sequence of Boolean formulae that need to be satisfied in order to complete the task.

\citet{vaezipoor2021LTL2Action} introduce LTL2Action, which directly encodes the LTL formula's syntax tree using a GNN and employs LTL progression~\citep{bacchus2000Using} to update the task representation. While this allows for generalization to some extent, the primary drawback of this method is that it requires the policy to \textit{learn} the semantics of temporal operators. Instead, we construct B\"uchi automata to explicitly capture the temporal structure of tasks, and condition the policy on sequences of simple \textit{Boolean} formulae.

Our work builds upon recent approaches that introduced the idea of exploiting the structure of B\"uchi automata for training a general LTL-conditioned policy~\citep{qiu2023Instructing,jackermeier2025deepltl}. However, our work differs in how the task is represented and processed. Instead of learning a policy conditioned directly on atomic propositions~\citep{qiu2023Instructing}, or on sequences of sets of assignments~\citep{jackermeier2025deepltl}, we propose to translate the transitions in the automaton into equivalent Boolean formulae and learn a policy conditioned on sequences thereof. This provides a more explicit and structured representation of the task transition dynamics, especially when multiple propositions can be simultaneously true and interact in complex ways, leading to better generalization and performance.

\section{Background}
\subsection{Reinforcement Learning}
We consider a standard reinforcement learning (RL) setup where an agent interacts with an environment, modeled as a Markov decision process (MDP). An MDP is defined as a tuple $\mathcal{M} = (\mathcal{S}, \mathcal{A}, p, r, \gamma, \rho_0)$, where $\mathcal{S}$ is the state space, $\mathcal{A}$ is the action space, $p: \mathcal{S} \times \mathcal{A} \times \mathcal{S} \rightarrow [0,1]$ is the transition probability function, $r: \mathcal{S} \times \mathcal{A} \times \mathcal{S} \rightarrow \mathbb{R}$ is the reward function, $\gamma \in [0,1)$ is the discount factor, and $\rho_0$ is the initial state distribution. The agent's goal is to learn a policy $\pi: \mathcal{S} \rightarrow \Delta(\mathcal{A})$ (a mapping from states to probability distributions over actions) that maximizes the expected discounted return $J(\pi) = \mathbb{E}_{\tau \sim \pi} [\sum_{t=0}^{\infty} \gamma^t r_t]$, where $\tau = (s_0, a_0, r_0, s_1, \dots)$ is a trajectory generated by following policy $\pi$ starting from $s_0 \sim \rho_0$, i.e., $a_t\sim\pi(\cdot\given s_t)$, $s_{t+1}\sim p(\cdot\given s_t,a_t)$, and $r_t = r(s_t,a_t,s_{t+1})$. The value function of a policy is defined as $V^\pi(s) = \mathbb{E}_{\tau \sim \pi} [\sum_{t=0}^{\infty} \gamma^t r_t\given s_0 = s]$, i.e., the expected discounted return of policy $\pi$ starting in state $s$.

\subsection{Linear Temporal Logic}
Linear temporal logic (LTL;~\citealp{pnueli1977temporal}) is a modal logic used to specify properties of infinite sequences of states, serving as a formal programmatic specification language. LTL formulae are defined over a set of atomic propositions ($AP$), which represent basic properties of the environment (e.g., "object $A$ is at location $X$").
The syntax of LTL is
$$ \varphi ::= \top \gror \mathsf p \gror \neg \varphi \gror \varphi \land \psi \gror \nex \varphi \gror \varphi \until \psi $$
where $\top$ denotes true, $\mathsf p \in AP$ is an atomic proposition, $\neg$~(negation) and $\land$ (conjunction) are standard Boolean connectives (from which others like $\lor$, $\rightarrow$, $\leftrightarrow$ can be derived), and $\mathsf X$ (next) and $\mathsf U$ (until) are temporal operators. Common derived temporal operators include $\event\varphi \equiv \top \until \varphi$ (eventually or finally) and $\always \varphi \equiv \neg \event \neg \phi$ (globally or always).

LTL semantics are defined over infinite traces $\sigma = \sigma_0 \sigma_1 \sigma_2 \dots$, where each $\sigma_t \subseteq \text{AP}$ is an \textit{assignment}, i.e., a set of atomic propositions true at time $t$. For an MDP, we assume a labeling function $L: \mathcal{S} \rightarrow 2^{AP}$ that maps each environment state $s \in \mathcal{S}$ to the set of atomic propositions true in that state. A trajectory $\tau = (s_0, a_0, r_0, s_1, \dots)$ satisfies an LTL formula $\varphi$, denoted $\tau \models \varphi$, if its corresponding trace $L(s_0)L(s_1)\dots$ satisfies $\varphi$ according to LTL semantics. For example, the LTL formula $\neg\mathsf a\until \mathsf b$ is satisfied by exactly the traces where eventually $\mathsf b$ is true at some time $t$, and $\mathsf a$ is false at all timesteps before. See \cref{app:ltl_semantics} for a formal definition of LTL satisfaction semantics. LTL provides a formal and structured way to define complex tasks to RL agents, such as "eventually reach region A, and if region B is entered, then eventually reach region C while avoiding region D."

\subsection{Büchi Automata for LTL}
The semantics of LTL formulae can be captured by Büchi automata~\citep{buchi1966Symposium}, which serve as \textit{explicit, programmatic structures} encoding a particular task. We here focus on limit-deterministic B\"uchi automata (LDBAs;~\citealp{sickert2016LimitDeterministic}), which are defined as tuples $\mathcal B = (\mathcal Q, q_0, \Sigma, \delta, \mathcal F, \mathcal E)$. $\mathcal Q = \mathcal Q_N \uplus \mathcal Q_D$ is a finite set of states partitioned into two subsets, $q_0\in\mathcal Q$ is the initial state, $\Sigma = 2^{AP}$ is a finite alphabet, $\delta\colon \mathcal Q\times(\Sigma\cup\mathcal E)\to \mathcal Q$ is the transition function, and $\mathcal F$ is the set of accepting states. We require that $\mathcal F\subseteq \mathcal Q_D$ and $\delta(q,\alpha)\in\mathcal Q_D$ for all $q\in\mathcal Q_D$ and $\alpha\in \Sigma$. The only way to transition from $\mathcal Q_N$ to $\mathcal Q_D$ is by taking a jump transition $\varepsilon\in\mathcal E$, which does not consume any input. Given an input trace $\sigma$, a \textit{run} of $\mathcal B$ is an infinite sequence of states in $\mathcal Q$ respecting the transition function $\delta$. A trace is \textit{accepted} by $\mathcal B$ if there exists a run that infinitely often visits accepting states.

\begin{theorem}[\citealp{sickert2016LimitDeterministic}]
    Given an LTL formula $\varphi$, it is possible to automatically construct an LDBA $\mathcal B_\varphi$ such that $\mathcal B_\varphi$ accepts exactly the traces $\sigma$ for which $\sigma \models\varphi$.
\end{theorem}

\begin{figure}
    \centering
        \resizebox{.65\columnwidth}{!}{
            \begin{tikzpicture}[->,>=stealth',shorten >=1pt,auto,semithick,]
                \node[state,initial] (q0) {$q_0$};
                \node[state] (q1) [above right=0.4cm and 1.4cm of q0] {$q_1$};
                \node[state,accepting] (qx) [right=1.2cm of q1] {$q_2$};
                \node[state,accepting] (q2) [below right=0.4cm and 1.4cm of q0] {$q_3$};
                \node[state] (q3) [right=1.2cm of q2] {$\bot$};
                \path (q0) edge [loop above] node {$\neg b$} (q0)
                      (q0) edge node {$b$} (q1)
                      (q1) edge node {$c$} (qx)
                      (q1) edge [loop above] node {$\neg c$} (q1) 
                      (qx) edge [loop above] node {$\top$} (qx) 
                      (q0) edge [below,pos=0.3] node {$\varepsilon_{q_3}$} (q2)
                      (q2) edge [loop above] node {$a$} (q2)
                      (q2) edge [above] node {$\neg a$} (q3)
                      (q3) edge [loop above] node {$\top$} (q3);
            \end{tikzpicture}
        }
        \caption{LDBA for the formula $(\event\always \mathsf a) \lor \event \mathsf (\mathsf b\land\event \mathsf c)$.}
        \label{fig:ldba}
\end{figure}

\begin{example}
    \cref{fig:ldba} shows an LDBA that accepts traces satisfying $(\event\always \mathsf a) \lor \event \mathsf (\mathsf b\land\event \mathsf c)$. Accepting runs either reach $q_1$ from $q_0$ via $\mathsf b$ and then $q_2$ via $\mathsf c$, or use the jump transition to reach $q_3$, after which $\mathsf a$ must hold true indefinitely.
\end{example}

B\"uchi automata are appealing structures to represent LTL instructions, since they explicitly capture the memory required to execute a given task. In RL settings, it is hence common to consider policies $\pi\colon \mathcal S\times\mathcal Q\to\Delta(\mathcal A)$ conditioned not only on the current MDP state $s$, but also on the current LDBA state $q$~\citep{hasanbeig2018LogicallyConstrained,hahn2019OmegaRegular}. After each environment interaction, the automaton state is updated according to the transition function $\delta$ and the currently true propositions $L(s)$. Over the course of a trajectory, updates to the LDBA state affect the behavior of the policy. For example, in \cref{fig:ldba} the policy initially might aim to make $\mathsf b$ true (following the upper branch), but once it has transition to state $q_1$, it would aim to make $\mathsf c$ true instead.

\begin{figure*}
    \centering
    \includegraphics[width=0.75\linewidth]{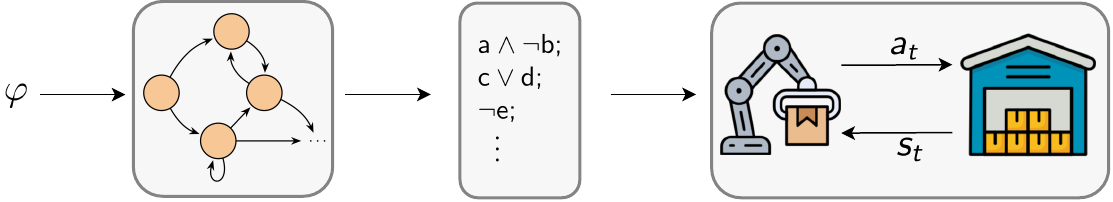}
    \caption{Given an LTL specification $\varphi$, we first construct an LDBA $\mathcal B_\varphi$ to represent the task structure and memory. We then extract a sequence of Boolean formulae from an accepting run in $\mathcal B_\varphi$ to condition our policy. If the LDBA state changes due to progress or external events, we recompute the accepting run from the new state $q'$ and re-condition the policy.}
    \label{fig:method-overview}
\end{figure*}

\section{Method}
We leverage goal-conditioned RL~\citep{liu2022GoalConditioned} to learn a generalist policy capable of executing arbitrary instructions specified in LTL\@. Our approach follows a standard framework~\citep{hasanbeig2018LogicallyConstrained,vaezipoor2021LTL2Action, jackermeier2025deepltl}: during training, we sample random LTL specifications $\varphi$ at the beginning of each episode, and generate a trajectory by repeatedly sampling $a_t\sim\pi(s_t\given \varphi)$. We simultaneously keep track of the current state of the LDBA $\mathcal B_\varphi$ constructed from $\varphi$, assigning positive rewards to actions that lead to accepting states in $\mathcal B_\varphi$, and giving a reward of 0 otherwise. Optimizing these rewards amounts to solving the following optimization problem:
\begin{equation*}
        \pi^* = \argmax_\pi \E_{\tau\sim\pi|\varphi}\left[
        \sum_{t=0}^\infty \gamma^t \bm 1_{\mathcal F_{\mathcal B_{\varphi}}}(q_t)]
        \right],
\end{equation*}
where $\bm 1_A$ is the indicator function of set $A$.
Intuitively, given an LTL instruction $\varphi$, the reward-optimal policy will visit accepting states in $\mathcal B_\varphi$ as often as possible, and hence satisfy the given specification. For a detailed discussion of how the reward-optimal policy relates to the optimal policy w.r.t.\ the probability of satisfying a given specification, see \citep{hahn2019OmegaRegular, voloshin2023Eventual,jackermeier2025deepltl}.

A key challenge in the above framework is how to condition the policy on a given LTL instruction $\varphi$. Prior work has shown that directly encoding $\varphi$, for example using a recurrent or graph neural network, tends to be ineffective, since this requires learning a complex non-stationary policy due to the non-Markovian nature of LTL tasks~\citep{vaezipoor2021LTL2Action}. Instead, we build on recent approaches that exploit the information contained in $\mathcal B_\varphi$ to learn a stationary policy $\pi\colon \mathcal S\times\mathcal Q\to\Delta(\mathcal A)$ conditioned on both the MDP state $s$ and current LDBA state $q$~\citep{qiu2023Instructing,jackermeier2025deepltl}.

\subsection{High-level Overview}
\cref{fig:method-overview} illustrates our approach. We first construct an LDBA $\mathcal B_\varphi$ from a given LTL instruction $\varphi$, which captures the task structure and the memory required to execute it. From the LDBA, we extract a sequence of \textit{Boolean formulae} corresponding to an accepting run from the initial state $q_0$. Each formula is associated with an edge in $\mathcal B_\varphi$, and succinctly represents the conditions that must hold true in order to make progress towards satisfying the task. We then execute a policy conditioned on this sequence of formulae in the environment. If the LDBA state changes to a new state $q'$, either because the policy has made progress towards the task or because of an external event, we recompute an accepting run from $q'$ and condition the policy on the corresponding new sequence of Boolean formulae.

\subsection{Extracting Accepting Runs}
Given an LDBA $\mathcal B_\varphi$ constructed from a specification $\varphi$, and a state $q$, we use Algorithm 1 of \cite{jackermeier2025deepltl} to identify accepting runs starting in $q$ (see \cref{app:alg}). This performs a simple depth-first search from $q$ to cycles in the automaton containing at least one accepting state. Each accepting run corresponds to one possible way of achieving the LTL instruction $\varphi$.

\subsection{Representing Runs via Boolean Formulae}
From an accepting run $\rho = (q, q_1, \ldots)$ we construct a sequence of Boolean formulae capturing the high-level goals the agent must achieve in order to follow the run, and hence satisfy the LTL task. Specifically, in order to transition from $q_i$ to $q_{i+1}$, the agent must achieve an assignment $a_i$ of atomic propositions that satisfies the transition condition $\delta(q_i, a_i) = q_{i+1}$, while avoiding any transitions to other states (excluding self-loops). We represent this with two Boolean formulae $\beta_i^+$ and $\beta_i^-$ that satisfy
\begin{align*}
    \forall a\in\mathbb A.\,a\models\beta_i^+ &\iff \delta(q_i,a) = q_{i+1},\\
    \forall a\in\mathbb A.\,a\models\beta_i^- &\iff \delta(q_i,a) \not\in \{q_i, q_{i+1}\},
\end{align*}
where $\mathbb A = \{L(s) : s\in\mathcal S\} \subseteq 2^{AP}$ is the set of possible assignments in the MDP\@. Intuitively, $\beta_i^+$ is a succint representation of the assignments that allow transitioning from $q_i$ to $q_{i+1}$, while $\beta_i^-$ captures the assignments that must not hold true in order to avoid transitioning to other states. 

\begin{example}
    Consider an MDP with propositions $AP = \{a,b,c,d\}$ in which all combinations of propositions can hold true at the same time. Assume we have an LDBA in which we can transition from state $q_0$ to $q_1$ via any of the assignments in the set
    \begin{equation*}
        A = \bigl\{ \{a\}, \{a, b\}, \{a, d\}, \{a, b, d\}\bigr\}.
    \end{equation*}
    A succinct representation of the transition from $q_0$ to $q_1$ is the formula $\beta_0^+ \equiv a\land\neg c$.
\end{example}

\paragraph{Constructing the formulae.}
To construct one of the formulae $\beta_i^+$ and $\beta_i^-$, we first identify the set of assignments $A$ for which it must hold true, i.e., the set of assignments corresponding to the relevant LDBA transitions. From this, we can trivially construct a disjunctive normal form (DNF) formula that captures the assignments in $A$. However, this representation is often not very informative, as it can be large and complex. Instead, we aim to find a small, semantically meaningful formula that captures the essence of the assignments in $A$ while enabling generalization at test time.

In general, this problem of finding a minimal Boolean formula for a given set of assignments is known to be intractable (assuming that $\mathsf{P} \neq \mathsf{NP}$)~\citep{masek1979some,allender2006minimizing}. We hence employ the following approximate procedure: we initially construct a set of well-structured candidate formulae by combining elementary \textit{formula templates} such as disjunctions of propositions, conjunctions of propositions, conjunctions with negated disjunctions, and combinations thereof. For each formula we construct, we compute its set of satisfying assignments $A\subseteq\mathbb A$, and finally for a given set of assignments we choose the associated candidate formula of minimal length. If we encounter a set of assignments for which no precomputed candidate formula exists, we fall back to using the DNF as described above. For further details on constructing candidate formulae, see \cref{app:formulae}.

\begin{figure*}[h!]
    \centering
    \includegraphics[width=.58\linewidth]{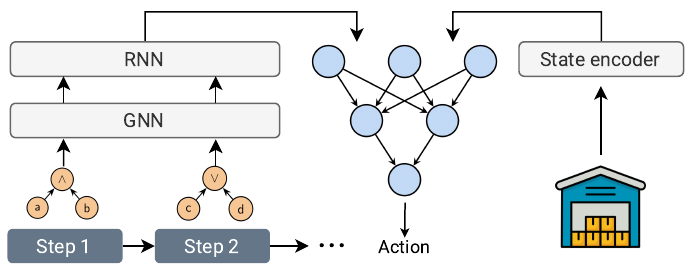}
    \caption{A sequence of Boolean formulae is processed via a GNN to produce a sequence of embeddings, which are passed through an RNN to obtain a representation of an accepting run. The MDP state is encoded via a state encoder. Both embeddings are fed into the policy network, which produces an action.}
    \label{fig:model}
\end{figure*}

\subsection{Learning Structured LTL Representations}
Having extracted sequences of Boolean formulae from the LDBA, we use a combination of GNNs and RNNs to obtain a meaningful task embedding for the policy. We first transform each formula into an abstract syntax tree (AST), where leaf nodes are propositions and internal nodes are logical operators. We interpret the AST as a directed graph with edges from children to parent nodes. Finally, we associate with each node in the AST a learnable embedding and apply a graph convolutional network (GCN;~\citealp{kipf2017Semisupervised}), which updates the node representations as follows:
\begin{equation*}
    \mathbf{h}_v^{(l+1)} = f\left( \sum_{u \in \mathcal{N}(v) \cup \{v\}} \frac{1}{\sqrt{d_v d_u}} \mathbf{W}^{(l)} \mathbf{h}_u^{(l)} \right),
\end{equation*}
where $\mathcal{N}(v)$ denotes the set of neighbors of node $v$, $d_v$ is the degree of node $v$ in the graph with added self-loops, $\mathbf{h}_v^{(l)}$ is the representation of node $v$ at layer $l$, $\mathbf{W}^{(l)}$ is a learnable weight matrix, and $f(\cdot)$ is a nonlinearity such as ReLU\@. The final embedding of the root node is the learned representation of the entire Boolean formula.

To obtain a meaningful representation of an accepting run $\rho$ in the LDBA, we concatenate the embeddings of $\beta_i^+$ and $\beta_i^-$ at each transition and produce an overall embedding by applying a gated recurrent neural network (GRU;~\citealp{cho2014Properties}). Since $\rho$ is generally an infinite sequence, we consider only a finite prefix of transitions as an approximation.


\paragraph{Policy Architecture.}
See \cref{fig:model} for an illustration of our overall policy architecture. Given an accepting run in the LDBA, we obtain an embedding from the GNN and RNN as discussed above. We simultaneously encode the current MDP state $s$ using either a multilayer perceptron (MLP) or convolutional neural network (CNN). The policy is instantiated as another MLP that maps from these embeddings to a distribution over actions, i.e., returns the parameters of a categorical or Gaussian distribution.

To handle jump transitions in the LDBA, we follow the standard procedure of augmenting the action space of the policy with designated $\varepsilon$-actions that take the jump transition without performing an action in the MDP~\citep{hasanbeig2018LogicallyConstrained,voloshin2023Eventual,jackermeier2025deepltl}. In the discrete case, we add an additional output logit to the policy network, and in the continuous case the policy network represents a mixture of a Gaussian distribution for the MDP action space, and a discrete distribution modelling the probability of taking an $\varepsilon$-action.

\subsection{Training}
We optimize the parameters of our model end-to-end via goal-conditioned RL\@. Instead of directly sampling LTL formulae during training, we design a training curriculum consisting of increasingly challenging sequences of Boolean formulae to satisfy. \citet{jackermeier2025deepltl} have previously shown that curriculum learning is an effective method for improving the training of LTL-conditioned policies in practice.

Let $\left\{ (\beta_i^+,\beta_i^-) \right\}_{i\in [n]}$ be a training sequence sampled from the curriculum. We assign a reward of 1 to an episode if the agent successfully satisfies the formulae $\beta_i^+$ in sequence, and assign a negative reward of $-1$ if the agent instead satisfies the currently active $\beta_i^-$. We jointly optimise the policy and learn a value function $V^\pi$ using proximal policy optimization~(PPO;~\citealp{schulman2017Proximal}).

\subsection{Selecting an Accepting Run}
In order to execute a new LTL instruction $\varphi$ with our trained model, we extract an accepting run of the B\"uchi automaton $\mathcal B_\varphi$ to condition the policy as described previously. However, in general there are multiple possible accepting runs for any given state of $\mathcal B_\varphi$. Similar to previous work~\citep{qiu2023Instructing,jackermeier2025deepltl} we use the \textit{value function} of the policy to select the best accepting run, i.e., the accepting run we are most likely to be able to complete and hence satisfy the task.

Specifically, let $s$ be the current MDP state, $q$ the current LDBA state, $AR = \{\rho_i\}_{i\in [n]}$ be the set of accepting runs computed via \cref{alg:cycles}, and $\varsigma$ be the function mapping a run to a sequence of Boolean formulae. We select the accepting run
\begin{equation*}
    \rho^* = \argmax_{\rho\in AR} V^\pi\left(s,\varsigma(\rho_i)\right)
\end{equation*}
to condition the policy each time the LDBA state is updated.

\subsection{Discussion}
Representing LTL instructions as sequences of Boolean formulae allows us to learn structured task embeddings that are useful for learning a generalist policy. In particular, Boolean formulae succinctly capture the \textit{meaning} of a \mbox{(sub-)}task and allow the policy to effectively generalize. This is especially true in settings where many atomic propositions can be true at the same time. For example, if the policy has learned how to achieve the formula $\mathsf a$ and how to achieve the formula $\mathsf b$, it can exploit what is has learned about the semantics of Boolean operators to also achieve the formula $\mathsf a\land\mathsf b$. Previous methods~\citep{qiu2023Instructing,jackermeier2025deepltl} do not support this type of generalization, and instead treat different assignments as completely separate goals for the policy.

\section{Experiments}
We conduct experiments to answer the following questions\footnote{Our code is available at \url{https://github.com/mattiagiuri/ltl_gnn}}: \textbf{(1)} Can our approach effectively zero-shot generalize to unseen LTL instructions? \textbf{(2)} How does our method compare to recent state-of-the-art approaches for training multi-task LTL-conditioned policies? \textbf{(3)} How does the performance of our method behave with increasing task difficulty?

\subsection{Experimental Setup}

\begin{figure}[t]
    \centering
    \includegraphics[width=.49\linewidth]{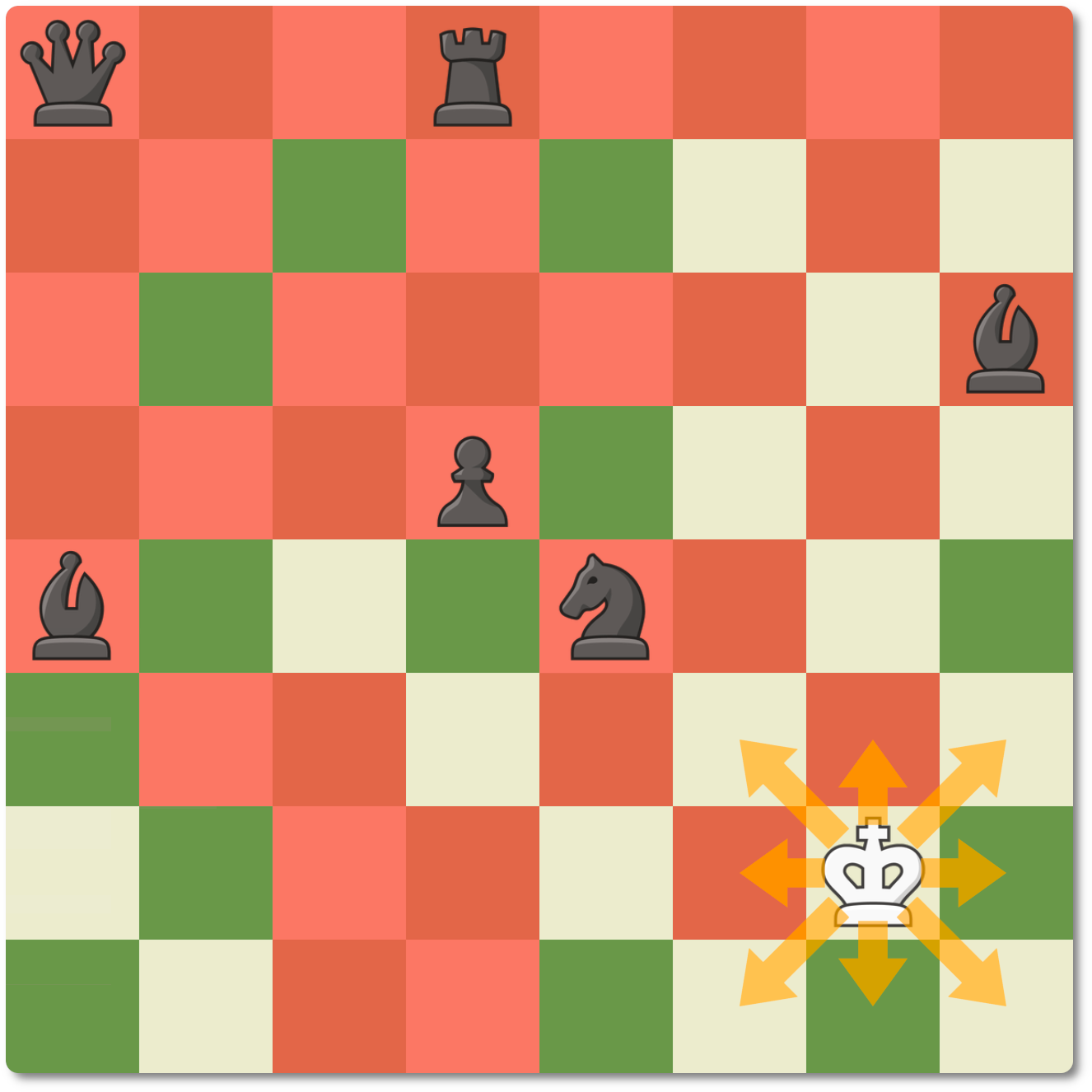}
    \caption{In the ChessWorld environment, the agent (the white king) must navigate the chessboard while avoiding certain squares attacked by black pieces. The agent can move along the 8 compass directions. Blue shading indicates squares attacked by at least one piece.}
    \label{fig:chessworld}
\end{figure}

\paragraph{ChessWorld.} 
We conduct our experiments in the ChessWorld environment, in which states correspond to positions on an $8\times 8$ chessboard. In the beginning of an episode, the agent\,---\,the white king\,---\,is randomly placed on an empty square. It then needs to navigate the board by moving along the 8 compass directions in order to reach squares that specific black pieces can move to, while avoiding squares attacked by other pieces. In particular, the atomic propositions correspond to the black pieces, i.e., $AP = \{\mathsf{queen}, \mathsf{rook}, \mathsf{knight}, \mathsf{bishop}, \mathsf{pawn}\}$, and a proposition is true in a square if the corresponding piece is either located on that square or attacks it. Due to the different movement patterns of the pieces, many squares can be attacked by multiple pieces at the same time, leading to a large number of possible assignments and complex interactions between propositions. For an illustration of the environment, see \cref{fig:chessworld}. More details about the environment can be found in \cref{app:chessworld}.

\paragraph{Tasks.}
We consider a number of tasks of varying difficulty to evaluate our method. We differentiate between \textit{finite-horizon} tasks, which can be solved in a finite number of steps, and \textit{infinite-horizon} tasks, which specify recurrent behavior that the agent must execute indefinitely. For example, the finite-horizon task $\event (\mathsf{queen} \land (\neg \mathsf{knight} \until \mathsf{rook}))$ requires the agent to reach a square attacked by the queen and subsequently avoid squares attacked by the knight until it reaches a square attacked by the rook. In contrast, the infinite-horizon task $\event \always \mathsf{queen}$ requires the agent to reach squares attacked by the queen and stay there indefinitely. We provide further details on the tasks used in our evaluation in \cref{app:tasks}.

\paragraph{Baselines.}
We compare our method to DeepLTL~\citep{jackermeier2025deepltl}, a state-of-the-art approach for learning a generalist policy for following LTL instructions. Similarly to our method, DeepLTL exploits the structure provided by B\"uchi automata for policy learning. However, it conditions the policy on sets of assignments without explicitly modelling their interactions. In contrast, our method represents the task as a sequence of Boolean formulae, which allows us to learn structured representations with GNNs.

We furthermore compare our approach to a novel baseline that augments DeepLTL with a Transformer encoder \citep{vaswani2017Attention} to learn meaningful representations of the sets of assignments. In theory, the attention mechanism of the Transformer should allow this model to learn interactions between the assignments, providing similar advantages as our approach.


    


    
    
    
    
    
    
    
    
    
    
    
    
    
    

\begin{table*}[ht]
    \centering
    \caption{Success rates (SR) and average discounted returns $J(\pi)$ on finite-horizon tasks with standard deviations over 5 seeds.\vspace{.2cm}}
    \label{tab:finite} 
    \begin{tabular}{@{}crrrrrr@{}}
    \toprule
    & \multicolumn{2}{c}{DeepLTL} & \multicolumn{2}{c}{Transformer} & \multicolumn{2}{c}{LTL-GNN} \\ 
    \cmidrule(lr){2-3} \cmidrule(lr){4-5} \cmidrule(lr){6-7}
    Task Set & SR & $J(\pi)$ & SR & $J(\pi)$ & SR & $J(\pi)$ \\
    \midrule
    $\phi_1$ & 99.1$_{\pm1.84}$ & 0.906$_{\pm0.017}$ & 70.6$_{\pm20.9}$ & 0.642$_{\pm0.187}$ & \textbf{99.3}$_{\pm0.92}$ & \textbf{0.915}$_{\pm0.009}$ \\
    $\phi_2$ & 92.3$_{\pm3.77}$ & 0.886$_{\pm0.037}$ & 57.6$_{\pm16.4}$ & 0.554$_{\pm0.159}$ & \textbf{95.2}$_{\pm3.24}$ & \textbf{0.913}$_{\pm0.029}$ \\
    $\phi_3$ & 68.8$_{\pm3.06}$ & 0.658$_{\pm0.030}$ & 51.7$_{\pm21.0}$ & 0.494$_{\pm0.200}$ & \textbf{82.6}$_{\pm6.60}$ & \textbf{0.785}$_{\pm0.061}$ \\
    $\phi_4$ & 91.2$_{\pm2.74}$ & 0.886$_{\pm0.027}$ & 77.0$_{\pm15.4}$ & 0.746$_{\pm0.153}$ & \textbf{92.7}$_{\pm3.87}$ & \textbf{0.902}$_{\pm0.036}$ \\
    $\phi_5$ & 67.1$_{\pm5.31}$ & 0.642$_{\pm0.050}$ & 41.7$_{\pm18.1}$ & 0.398$_{\pm0.171}$ & \textbf{74.3}$_{\pm8.52}$ & \textbf{0.709}$_{\pm0.078}$ \\
    $\phi_6$ & 91.5$_{\pm2.32}$ & 0.873$_{\pm0.023}$ & 76.6$_{\pm17.1}$ & 0.724$_{\pm0.162}$ & \textbf{93.6}$_{\pm1.95}$ & \textbf{0.892}$_{\pm0.016}$ \\
    $\phi_7$ & \textbf{91.9}$_{\pm1.34}$ & \textbf{0.897}$_{\pm0.014}$ & 80.8$_{\pm11.1}$ & 0.785$_{\pm0.110}$ & 91.0$_{\pm3.24}$ & 0.888$_{\pm0.031}$ \\
    \bottomrule
    \end{tabular}
\end{table*}

\begin{table}[ht]
    \centering
    \caption{Success rates on infinite-horizon tasks with standard deviations over 5 seeds.\vspace{.2cm}}
    \label{tab:infinite}

    \begin{tabular}{@{}lrrr@{}}
        \toprule
        Task Set & \multicolumn{1}{c}{DeepLTL} & \multicolumn{1}{c}{Transformer} & \multicolumn{1}{c}{LTL-GNN} \\
        \midrule
        $\phi^{\infty}_{\text{GF}}$ & \textbf{95.7}$_{\pm01.7}$ & 67.6$_{\pm28.5}$ & 92.8$_{\pm9.3}$ \\
        $\phi^{\infty}_1$       & 40.0$_{\pm49.0}$ & 43.1$_{\pm42.2}$ & \textbf{86.0}$_{\pm28.0}$ \\
        $\phi^{\infty}_2$            & 35.7$_{\pm44.0}$ & 33.6$_{\pm33.9}$ & \textbf{76.7}$_{\pm36.7}$ \\
        \bottomrule
    \end{tabular}
\end{table}

\paragraph {Evaluation protocol.}
All methods are trained with the same training curriculum (see \cref{app:curriculum}) for 15M interaction steps. Hyperparameter details are provided in \cref{app:hyperparameters}. We report performance in terms of success rate (SR) and average discounted return $J(\pi)$. All results are averaged over 5 random seeds.

\subsection{Results}
\begin{figure*}[t]
\centering
\begin{minipage}[t]{.49\textwidth}
    \centering
    \includegraphics[width=0.8\linewidth]{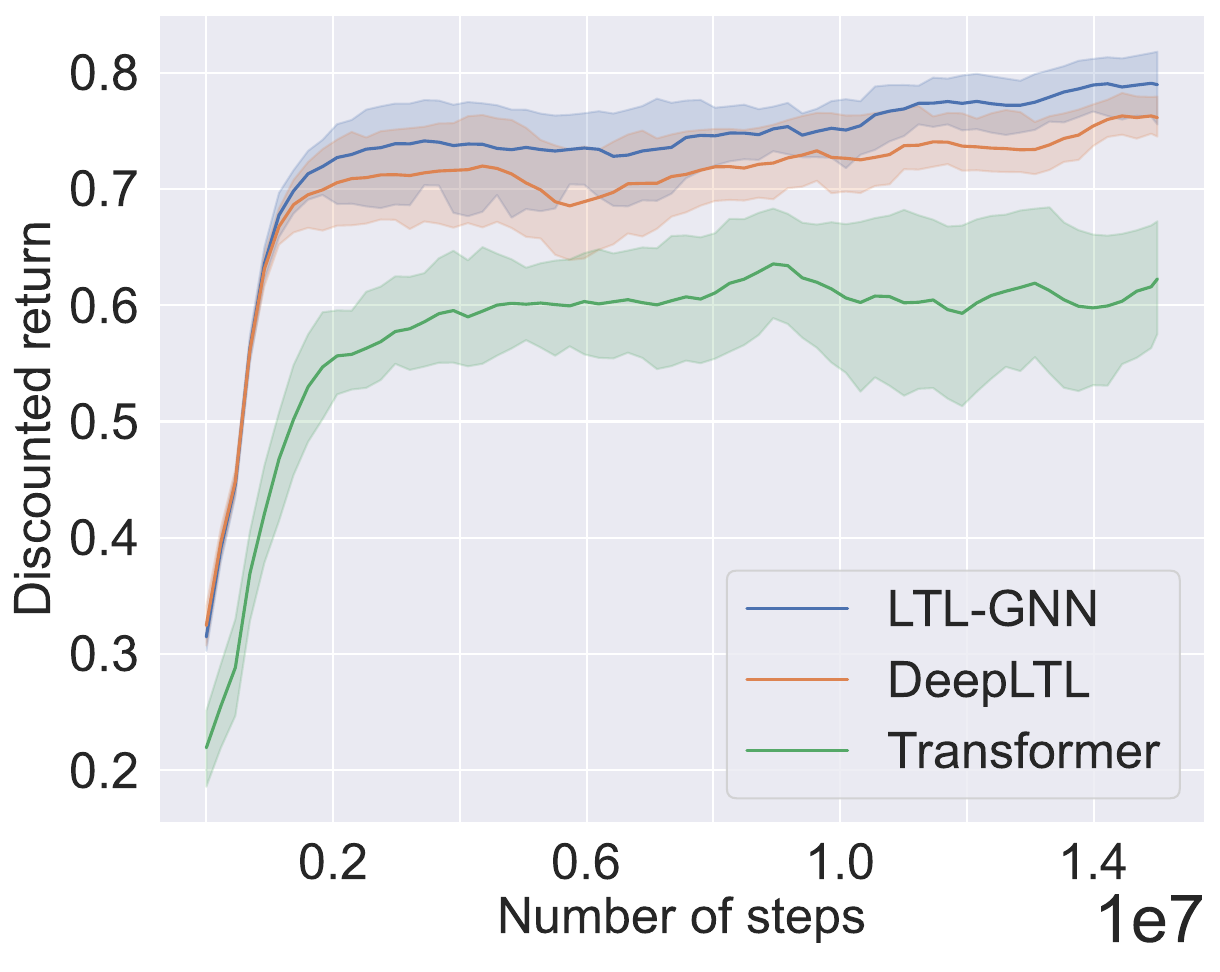}
\caption{Average discounted returns $J(\pi)$ over the number of interaction steps during training. Shaded areas indicate 90\% confidence intervals over 5 random seeds.}
\label{fig:curves-training}
\end{minipage}\hfill
\begin{minipage}[t]{.49\textwidth}
    \centering
    \includegraphics[width=0.8\linewidth]{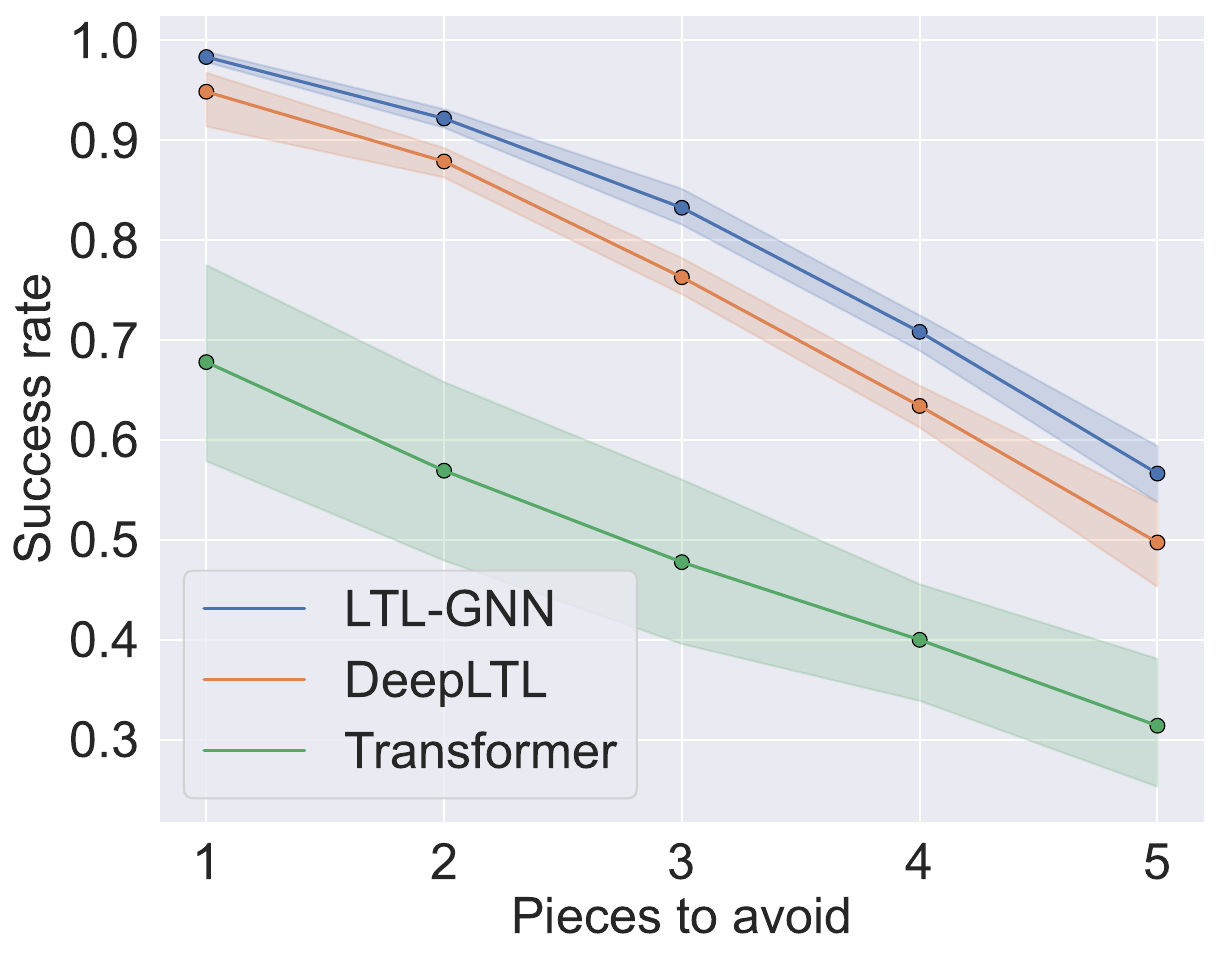}
\caption{Success rates over tasks with increasing difficulty in terms of the number of pieces to avoid. Shaded areas indicate 90\% confidence intervals over 5 random seeds.}
\label{fig:pieces-ablation}
\end{minipage}
\end{figure*}

\cref{tab:finite,tab:infinite} show the results of evaluating the trained policies on finite-horizon and infinite-horizon tasks, respectively. We furhermore show the discounted return on finite-horizon tasks over training in \cref{fig:curves-training}. We see that our method, denoted as LTL-GNN, is able to effectively generalize to unseen LTL instructions and significantly outperforms the baselines on most tasks, achieving higher success rates and discounted returns. This is particularly evident in the challenging infinite-horizon tasks, where both baselines fail to learn a policy that can consistently satisfy tasks in $\phi_1^\infty$ and $\phi_2^\infty$.

The results demonstrate the advantages of our structured learned task representations based on sequences of Boolean formulae. In particular, the GNN-based representation allows us to learn meaningful task embeddings that capture the interactions between the assignments in the LTL tasks. This is in contrast to the Transformer baseline, which struggles to learn effective representations of the sets of assignments.

\paragraph{Varying the Task Difficulty.}
We further evaluate the performance of our method and the baselines on tasks of varying difficulty, in which we increase the number of pieces that the agent must avoid in order to complete a task. We consider reach-avoid tasks of the form $\neg\mathsf a\until\mathsf b$ and vary the number of pieces that must be avoided from 1 to 5. The results are shown in \cref{fig:pieces-ablation}.

As expected, we see that the success rates of all methods decrease with increasing task difficulty, i.e., as the number of pieces to avoid increases. However, our method is able to maintain higher success rates than the baselines even for the most difficult tasks, demonstrating its strong performance on challenging LTL instructions.

\section{Conclusion and Future Work}

We have presented a novel approach for learning generalist policies for following LTL tasks based on structured task representations. Our method exploits the structure of B\"uchi automata constructed from a given specification, and extracts sequences of Boolean formulae that succinctly represent different ways of achieving the task. These formulae are encoded by a combination of GNNs and RNNs to condition the policy on the LTL instruction. In contrast to previous methods, this allows us to effectively model interactions between different assignments, and to generalize from simpler to more complicated tasks. We have shown that our approach is able to effectively zero-shot generalize to unseen LTL instructions and outperforms state-of-the-art methods in terms of success rate and discounted return.

There are several interesting directions for future work. It would be interesting to apply our method to larger, more realistic environments with high-dimensional observation and action spaces. In some environments, especially vision-based, the labeling function mapping from observations to atomic propositions may not be known. Future work could explore jointly learning the labeling function or incorporating pre-trained foundation models as high-level event detectors. Generally, these advances could lead to better generalist models capable of following well-defined instructions.

\bibliography{rl,mc,ml}
\bibliographystyle{icml2025}

\newpage
\appendix
\onecolumn
\section{Semantics of LTL}
\label{app:ltl_semantics}

The satisfaction semantics of LTL are specified by the satisfaction relation $\sigma\models\varphi$, which is recursively defined as follows~\citep{baier2008Principles}:
\begin{align*}
    \sigma &\models \top\\
    \sigma &\models \mathsf a &&\text{iff } \mathsf a\in \sigma_0 \\
    \sigma &\models \varphi\land\psi &&\text{iff } \sigma\models\varphi \land \sigma\models\psi \\
    \sigma &\models \neg\varphi &&\text{iff } \sigma\not\models\varphi \\
    \sigma &\models \nex\varphi &&\text{iff } \sigma[1\ldots]\models\varphi \\
    \sigma &\models \varphi\;\mathsf{U}\;\psi &&\text{iff } \exists j\geq 0.\; \sigma[j\ldots]\models\psi \land \forall 0\leq i < j.\; \sigma[i\ldots]\models\varphi.
\end{align*}

\section{Algorithm to Identify Accepting Runs}
\label{app:alg}

\begin{algorithm}
	\caption{Computing paths to accepting cycles~\citep{jackermeier2025deepltl}}
	\label{alg:cycles}
	\begin{algorithmic}[1]
		\Require
			\Statex An LDBA $B = (\mathcal Q, q_0, \Sigma, \delta, \mathcal F, \mathcal E)$ and current state $q$.
		\Procedure{DFS}{$q$, $p$, $i$}\Comment{$i$ is in the index of the last seen accepting state, or $-1$ otherwise}
        \State $P\gets \emptyset$
        \If{$q\in\mathcal F$}
            \State $i\gets |p|$
        \EndIf
        \ForAll {$a \in 2^{AP}\cup\{\varepsilon\}$}
            \State $p'\gets [p,q]$
            \State $q' \gets \delta(q, a)$
            \If{$q'\in p$}
                \If{index of $q'$ in $p$ $\leq i$}
                    \State $P = P \cup \{p'\}$
                \EndIf
            \Else
                \State $P = P\,\cup$ \textsc{DFS}($q'$, $p'$, $i$)
            \EndIf
        \EndFor
        \State \textbf{return} $P$
		
		\EndProcedure
        \State $i\gets 0 $ if $q\in\mathcal F $ else $i\gets -1$
        \State \textbf{return } \textsc{DFS}($q$, $[]$, $i$)
	\end{algorithmic}
\end{algorithm}


\section{Experimental Details}

\subsection{ChessWorld}
\label{app:chessworld}

The ChessWorld environment cosists of an $8\times 8$ chessboard. Each state $(x,y)\in [8]^2$ corresponds to a square on the board. The atomic propositions correspond to the black pieces on the chessboard. If piece $p$ is located on square $s = (x,y)$ or $p$ attacks square $s$, then $p$ is active in $s$, i.e., $p\in L(s)$. \cref{tab:chess-assignments} lists all combinations of propositions that hold true at some state in ChessWorld. The action space consists of 9 possible actions: the possible 8 moves of a king in chess (i.e., move one square orthogonally or diagonally), and a ``stay" action that does nothing.

\begin{table}[h!]
\centering
\caption{Possible assignments in ChessWorld.\vspace{0.2cm}}

\begin{tabular}{@{}ccc@{}}
        \toprule
        $\{\mathsf{queen}\}$ & $\{\mathsf{rook}\}$ & $\{\mathsf{knight}\}$ \\
        $\{\mathsf{bishop}\}$ & $\{\mathsf{pawn}\}$ & $\{\mathsf{queen}, \mathsf{rook}\}$ \\
        $\{\mathsf{queen}, \mathsf{bishop}\}$ & $\{\mathsf{queen}, \mathsf{pawn}, \mathsf{bishop}\}$ & $\{\mathsf{queen}, \mathsf{pawn}, \mathsf{rook}\}$ \\
        $\{\mathsf{knight}, \mathsf{rook}\}$ & $\{\mathsf{bishop}, \mathsf{rook}\}$ & $\{\mathsf{knight}, \mathsf{bishop}\}$ \\
        \bottomrule
    \end{tabular}
\label{tab:chess-assignments}
\end{table}





\subsection{Hyperparameters}
\label{app:hyperparameters}
\paragraph{Neural Networks.} In all experiments, our policy is instantiated with a fully connected neural network with dimensions of [128, 64, 64] and ReLU activations. Its output is a categorical distribution, modeled by a softmax layer. The critic network has [128, 64] units with ReLU activations. 

We employ a GCN with hidden dimension $d=32$ and 3 layers. The DeepSets unit in DeepLTL uses a [32, 32] feed-forward network with ReLU activations. The Transformer encoder for the baseline has the following structure: pre-layer normalization, multi-head self-attention with 2 heads and dimension 32, residual connection and layer normalization, residual connection and a [32, 32] feed-forward network.
\begin{table}[]
    \centering
    \caption{PPO hyperparameters.\vspace{.2cm}}
    \label{tab:ppo_params}
    \begin{tabular}{lr}
        \toprule
        Parameter & Value \\
        \midrule
        Number of processes & 16 \\
        Steps per process per update & 2048 \\
        Epochs & 10 \\
        Batch size & 4096 \\
        Discount factor & 0.98 \\
        GAE-$\lambda$ & 0.95 \\
        Entropy coefficient & 0.003 \\
        Value loss coefficient & 0.5 \\
        Max gradient norm & 0.5 \\
        Clipping ($\epsilon$) & 0.2 \\
        Adam learning rate & 0.0003 \\
        Adam epsilon & 1e-08 \\
        \bottomrule
    \end{tabular}
\end{table}

\textbf{PPO.} The hyperparameters for PPO \cite{schulman2017Proximal} are listed in \cref{tab:ppo_params}. We use Adam~\citep{kingma2015Adam} for all experiments.

\subsection{Training Curriculum}
\label{app:curriculum}

The curriculum consists of three stages, each characterised by a distribution over sequences of Boolean formulae with task difficulty increasing over time. Tasks involve logical combinations (i.e., conjunctions/disjunctions) of atomic propositions, e.g., $\mathsf{bishop} \land \mathsf{queen}$, $\mathsf{rook} \lor \mathsf{pawn}$, or $\mathsf{knight} \land \lnot \mathsf{bishop}$. In the first stage, the focus is on simple reach-only tasks and reach-avoid tasks with at most one piece to avoid. In the second stage, the avoid formulae expand to include up to three pieces, and we introduce more complex finite tasks (e.g., "avoid being attacked until reaching a piece") as well as reach-stay specifications. The third stage further increases the challenge by requiring longer persistence in reach-stay tasks, while keeping the same finite-horizon tasks as the previous stage. For DeepLTL and the Transformer baseline we translate the sequence of Boolean formulae back to a sequence of assignments.

\subsection{Experiment Tasks}
\label{app:tasks}
\cref{tab:finite-horizon-tasks,tab:infinite-horizon-tasks} list the LTL tasks used in our evaluation.

\begin{table}[h!]
\centering
\caption{ChessWorld finite-horizon LTL specifications.\vspace{.2cm}}
\label{tab:finite-horizon-tasks}
\begin{tabular}{lp{11cm}}
\toprule
Task Set & Specifications \\ \midrule
$\phi_1$ &
\begin{tabular}[t]{@{}l@{}}
$\event (\mathsf{pawn} \land \event (\mathsf{rook} \land \event \mathsf{knight}))$ \\
$\event ((\mathsf{rook} \land \mathsf{queen}) \land \event \mathsf{bishop})$ \\
$\event (\mathsf{bishop} \land \mathsf{rook}) \land \event (\mathsf{bishop} \land \mathsf{knight})$
\end{tabular} \\ \midrule
$\phi_2$ &
\begin{tabular}[t]{@{}l@{}}
$\neg (\mathsf{pawn} \lor \mathsf{bishop}) \until (\mathsf{bishop} \land \mathsf{rook})$ \\
$\neg (\mathsf{queen} \lor \mathsf{pawn}) \until (\mathsf{rook} \land \mathsf{queen})$ \\
$\neg (\mathsf{bishop} \lor \mathsf{pawn}) \until (\mathsf{rook} \land \mathsf{knight})$ \\
$\neg (\mathsf{knight} \lor \mathsf{rook}) \until \mathsf{bishop}$ \\
$\neg (\mathsf{bishop} \lor \mathsf{knight}) \until \mathsf{queen}$ \\
$\neg (\mathsf{rook} \lor \mathsf{bishop}) \until \mathsf{pawn}$
\end{tabular} \\ \midrule
$\phi_3$ &
\begin{tabular}[t]{@{}l@{}}
$\neg (\mathsf{bishop} \lor \mathsf{knight} \lor \mathsf{pawn}) \until (\mathsf{rook} \land \mathsf{queen})$ \\
$\neg (\mathsf{knight} \lor \mathsf{rook} \lor \mathsf{bishop}) \until (\mathsf{rook} \land \mathsf{bishop})$ \\
$\neg (\mathsf{bishop} \lor \mathsf{pawn} \lor \mathsf{rook}) \until (\mathsf{rook} \land \mathsf{queen})$ \\
$\neg (\mathsf{bishop} \lor \mathsf{knight} \lor \mathsf{queen}) \until (\mathsf{rook} \land \mathsf{queen})$
\end{tabular} \\ \midrule
$\phi_4$ &
\begin{tabular}[t]{@{}l@{}}
$\neg (\mathsf{bishop} \lor \mathsf{rook} \lor \mathsf{knight} \lor \mathsf{pawn}) \until \mathsf{queen}$ \\
$\neg (\mathsf{bishop} \lor \mathsf{rook} \lor \mathsf{knight} \lor \mathsf{queen}) \until \mathsf{pawn}$ \\
$\neg (\mathsf{bishop} \lor \mathsf{rook} \lor \mathsf{pawn} \lor \mathsf{queen}) \until \mathsf{knight}$ \\
$\neg (\mathsf{bishop} \lor \mathsf{knight} \lor \mathsf{pawn} \lor \mathsf{queen}) \until \mathsf{rook}$ \\
$\neg (\mathsf{rook} \lor \mathsf{knight} \lor \mathsf{pawn} \lor \mathsf{queen}) \until \mathsf{bishop}$
\end{tabular} \\ \midrule
$\phi_5$ &
\begin{tabular}[t]{@{}l@{}}
$\neg (\mathsf{bishop} \lor \mathsf{rook} \lor \mathsf{knight} \lor \mathsf{pawn} \lor \mathsf{queen}) \until (\mathsf{queen} \land \mathsf{pawn})$ \\
$\neg (\mathsf{bishop} \lor \mathsf{rook} \lor \mathsf{knight} \lor \mathsf{queen} \lor \mathsf{pawn}) \until (\mathsf{pawn} \land \mathsf{rook})$ \\
$\neg (\mathsf{bishop} \lor \mathsf{rook} \lor \mathsf{pawn} \lor \mathsf{queen} \lor \mathsf{knight}) \until (\mathsf{knight} \land \mathsf{bishop})$ \\
$\neg (\mathsf{bishop} \lor \mathsf{knight} \lor \mathsf{pawn} \lor \mathsf{queen} \lor \mathsf{rook}) \until (\mathsf{rook} \land \mathsf{knight})$ \\
$\neg (\mathsf{rook} \lor \mathsf{knight} \lor \mathsf{pawn} \lor \mathsf{queen} \lor \mathsf{bishop}) \until (\mathsf{bishop} \land \mathsf{queen})$ \\
$\neg (\mathsf{rook} \lor \mathsf{knight} \lor \mathsf{pawn} \lor \mathsf{queen} \lor \mathsf{bishop}) \until (\mathsf{rook} \land \mathsf{queen})$
\end{tabular} \\ \midrule
$\phi_6$ &
\begin{tabular}[t]{@{}l@{}}
$\event (\mathsf{queen} \land (\neg \mathsf{knight} \until \mathsf{rook}))$ \\
$\neg (\mathsf{pawn} \lor \mathsf{knight}) \until (\mathsf{queen} \land \mathsf{rook}) \land \event \mathsf{pawn}$ \\
$\neg (\mathsf{bishop} \lor \mathsf{rook}) \until \mathsf{pawn} \land \event \mathsf{knight}$ \\
$\event (\mathsf{rook} \land (\neg \mathsf{bishop} \until \mathsf{pawn}))$ \\
$(\neg \mathsf{queen} \until \mathsf{pawn}) \land (\neg \mathsf{bishop} \until \mathsf{knight})$ \\
$(\neg \mathsf{queen} \until \mathsf{rook}) \land (\neg \mathsf{knight} \until \mathsf{queen})$ \\
$(\neg \mathsf{queen} \until \mathsf{pawn}) \land (\neg \mathsf{bishop} \until \mathsf{knight}) \land (\neg \mathsf{knight} \until \mathsf{rook})$
\end{tabular} \\ \midrule
$\phi_7$ &
\begin{tabular}[t]{@{}l@{}}
$\neg (\mathsf{rook} \lor \mathsf{bishop} \lor \mathsf{pawn}) \until (\mathsf{knight} \land \neg \mathsf{rook})$ \\
$\neg \mathsf{queen} \until (\mathsf{bishop} \land \neg \mathsf{pawn})$ \\
$\neg (\mathsf{bishop} \lor \mathsf{knight}) \until (\mathsf{queen} \land \neg \mathsf{knight})$ \\
$\neg (\mathsf{rook} \lor \mathsf{knight} \lor \mathsf{queen} \lor \mathsf{pawn}) \until (\mathsf{bishop} \land \neg \mathsf{queen})$ \\
$\neg (\mathsf{pawn} \lor \mathsf{queen} \lor \mathsf{rook} \lor \mathsf{knight} \lor \mathsf{bishop}) \until (\mathsf{rook} \land \neg \mathsf{bishop})$
\end{tabular} \\ \bottomrule
\end{tabular}
\end{table}

\begin{table}[h!]
\centering
\caption{ChessWorld infinite-horizon LTL specifications.\vspace{.2cm}}
\label{tab:infinite-horizon-tasks}
\begin{tabular}{lp{9cm}}
\toprule
Task Set & Specifications \\ 
\midrule

$\phi^{\infty}_{GF}$ &
\begin{tabular}[t]{@{}l@{}}
$\always \event \mathsf{knight} \land \always \event \mathsf{queen}$ \\
$\always \event \mathsf{pawn} \land \always \event \mathsf{rook}$ \\
$\always \event \mathsf{bishop} \land \always \event \mathsf{knight} \land \always \neg \mathsf{rook}$ \\
$\always \event \mathsf{rook} \land \always \event \mathsf{pawn} \land \always \neg \mathsf{knight}$
\end{tabular} \\ \midrule
$\phi^{\infty}_1$ &
\begin{tabular}[t]{@{}l@{}}
$\event \always \mathsf{bishop}$ \\
$\event \always \mathsf{queen}$ \\
$\event \always \mathsf{rook}$ \\
$\event \always \mathsf{pawn}$ \\
$\event \always \mathsf{knight}$ \\
$\event \always (\mathsf{queen} \lor \mathsf{bishop})$ \\
$\event \always (\mathsf{rook} \lor \mathsf{queen})$ \\
$\event \always (\mathsf{knight} \lor \mathsf{pawn})$ \\
$\event \always (\mathsf{bishop} \lor \mathsf{knight})$ \\
$\event \always (\mathsf{rook} \lor \mathsf{pawn})$
\end{tabular} \\ \midrule
$\phi^{\infty}_2$ &
\begin{tabular}[t]{@{}l@{}}
$\event \always (\mathsf{bishop} \land \neg \mathsf{rook})$ \\
$\event \always (\mathsf{knight} \land \neg \mathsf{bishop})$ \\
$\event \always (\mathsf{queen} \land \mathsf{pawn})$ \\
$\event \always (\mathsf{rook} \land \mathsf{queen})$
\end{tabular} \\ \bottomrule
\end{tabular}
\end{table}

\section{Mapping Sets of Assignments to Boolean Formulae}
\label{app:formulae}

The construction of the mapping begins with a curated dataset of desirable formulae, denoted as $FD$. Each formula $\psi \in FD$ is translated into a set of assignments $G_\psi$, using a mapping that replaces logical operators with set-theoretic counterparts (e.g. $\land \to \cap$). The collection $FA = \{G_\psi : \psi \in FD\}$ constitutes the assignment space corresponding to the formula dataset.

An injective mapping $FC: FA \to FD$ is then established, ensuring each assignment set is uniquely linked to an optimal formula under a chosen metric. The mapping is designed to include typical LTL-derived assignments; unseen sets during deployment are processed by generating their DNF.

\subsection{Notation and Conventions}
\label{sec:cache-conventions}

Let the set of environment variables be denoted by $V$. We define $P(V)$ as the power set of $V$, excluding the empty set and $V$ itself. For fixed cardinality, $P(V)_k$ refers to the subset of $P(V)$ containing only sets of size $k$, i.e., $\binom{V}{k}$. More generally, we define $P(V)_x^y$ as the collection of sets of size $y$ drawn from $P(V)_x$, i.e., $\binom{P(V)_x}{y}$. Each element in $P(V)_x^y$ is thus a set of $y$ subsets of $V$, each of cardinality $x$.

The mapping is built by composing unions, intersections, and differences of sets $G_a = \set{\alpha \,|\, \alpha \in \mathbb A, a \in \alpha}$, where $a \in AP$. It is implemented as a dictionary that maps sets of assignments to logical formulae. The core strategy is to define and insert a curated sequence of \textit{templates}—canonical forms of formulae—into the mapping. Since multiple formulae $\psi_i$ may represent the same set of assignments $G$, the insertion order of templates reflects increasing syntactic complexity, ensuring that $G$ is represented by the simplest possible formula.

Importantly, for every formula added to the mapping, a corresponding complement formula (representing $\mathbb A\setminus G$) is also inserted. Throughout, $\text{range}(a, b)$ denotes the set of integers in $[a, b{-}1]$. Each of the following subsections defines a specific \textit{template} in the order in which they are added to the mapping.

\subsection{No Assignments}
The empty set of assignments is added at the start, in order to prevent edge cases (i.e., formulae representing empty assignment sets are not added).


\subsection{\textit{Or} Formulae}
These formulae are of the form
$$\bigvee_{v \in \mathcal{V}} v \quad \forall \; \mathcal{V} \in P(V)_k, \quad k=1, 2, .., |V|-1$$

\noindent with corresponding complement formulae
$$\neg \brac{\bigvee_{v \in \mathcal{V}} v} \quad \forall \; \mathcal{V} \in P(V)_k, \quad k=1, 2, .., |V|-1$$

\noindent These are added to the mapping in increasing order of $k$, i.e. for $k=1, 2, .., |V|-1$.

\subsection{\textit{And} Formulae}
\label{sec:and-formulae}

These formulae are of the form
$$\bigwedge_{v \in \mathcal{V}} v \quad \forall \; \mathcal{V} \in P(V)_k, \quad k=1, 2, .., |V|-1$$

\noindent with corresponding complement formulae
$$\neg \brac{\bigwedge_{v \in \mathcal{V}} v} \quad \forall \; \mathcal{V} \in P(V)_k, \quad k =1, 2, .., |V|-1$$

\noindent These are added to the mapping in increasing order of $k$, i.e. for $k=1, 2, .., |V|-1$.

\subsection{\textit{Or-x-and-y} Formulae}
Here $x$ and $y$ are parameters to be chosen, in our experiments $x = 4, y = 2$. The formulae are
$$\brac{\bigvee_{v \in \mathcal{V}_1} v} \land \brac{\bigwedge_{w \in \mathcal{V}_2}w} \quad \begin{cases}\forall \; \mathcal{V}_1 \in P(V)_i, &\text{for } i \in \text{range}(2, x+1) \\ \forall \; \mathcal{V}_2 \in P(V)_j \text{ st } \mathcal{V}_2 \cap \mathcal{V}_1 = \emptyset, & \text{for } j \in \text{range}(1, y+1),
\end{cases}$$
 added to the mapping in lexicographical order of $(i, j)$. The complement formulae are
$$\neg \sbrac{\brac{\bigvee_{v \in V_i} v} \land \brac{\bigwedge_{w \in V_j}w}}.$$

\subsection{\textit{And-x-and-not-y} Formulae}
Here $x$ and $y$ are parameters to be chosen, in our experiments $x = 2, y = 3$. The formulae are
$$\brac{\bigwedge_{v \in \mathcal{V}_1} v} \land \neg \brac{\bigvee_{w \in \mathcal{V}_2}w} \quad \begin{cases}\forall \; \mathcal{V}_1 \in P(V)_i, &\text{for } i \in \text{range}(2, x+1) \\ \forall \; \mathcal{V}_2 \in P(V)_j \text{ st } \mathcal{V}_2 \cap \mathcal{V}_1 = \emptyset, & \text{for } j \in \text{range}(1, y+1),
\end{cases}$$
added to the mapping in lexicographical order of $(i, j)$. The complement formulae are
$$\neg\brac{\bigwedge_{v \in \mathcal{V}_1} v} \lor \brac{\bigvee_{w \in \mathcal{V}_2}w}$$

\subsection{\textit{Or-x-and-not-y} Formulae}
Here $x$ and $y$ are parameters to be chosen, in our experiments $x = 4, y = 4$. The formulae are
$$\brac{\bigvee_{v \in \mathcal{V}_1} v} \land \neg \brac{\bigvee_{w \in \mathcal{V}_2}w} \quad \begin{cases}\forall \; \mathcal{V}_1 \in P(V)_i, &\text{for } i \in \text{range}(1, x+1) \\ \forall \; \mathcal{V}_2 \in P(V)_j \text{ st } \mathcal{V}_2 \cap \mathcal{V}_1 = \emptyset, & \text{for } j \in \text{range}(1, y+1),
\end{cases}$$
added to the mapping in lexicographical order of $(i, j)$. The complement formulae have the form
$$\neg \brac{\bigvee_{v \in \mathcal{V}_1} v} \lor \brac{\bigvee_{w \in \mathcal{V}_2}w}$$

\subsection{\textit{Or-x-and-not-zy} Formulae}
Here $x$, $y$ and$ $z are parameters to be chosen, in our experiments $x=4, y=3, z=2$. 
$$\brac{\bigvee_{v \in \mathcal{V}_1} v} \land \neg \sbrac{\bigvee_{\mathcal{V}_2 \in P(V)_j^k} \brac{\bigwedge_{w \in \mathcal{V}_2} w}} \quad \begin{cases}\forall \; \mathcal{V}_1 \in P(V)_i, & \text{for } i \in  \text{range}(1, x+1) \\ \forall \; \mathcal{V}_2 \in P(V)_j^k, & \text{for } j \in \text{range}(2, y+1), k \in \text{range}(1, z+1),
\end{cases}$$
added to the mapping in lexicographical order of $(i, j, k)$. The complement formulae are
$$\neg \brac{\bigvee_{v \in \mathcal{V}_1} v} \lor  \sbrac{\bigvee_{\mathcal{V}_2 \in P(V)_j^k} \brac{\bigwedge_{w \in \mathcal{V}_2} w}}.$$

\subsection{Choosing a Formula}
In general there may be sets of assignments with multiple matching formulae, that is, the mapping $FC$ need not be injective. In this case, we select the first matching formula added to the mapping, which implicitly corresponds to selecting the minimal formula in terms of number of operators, i.e., we select
$$\psi^*_G = \arg \min_{\psi \in F_G} c(\psi),$$
where $c(\psi)$ measures the complexity of formula $\psi$.

\end{document}